\documentclass[runningheads]{llncs}
\usepackage[T1]{fontenc}
\usepackage{graphicx}
\usepackage{multirow}
\usepackage{tabularx}
\usepackage{amsmath}
\usepackage{amsxtra}
\usepackage{amsfonts}
\usepackage{multicol}
\usepackage{amssymb}
\usepackage{booktabs}
\usepackage[table,xcdraw]{xcolor}
\usepackage{subcaption}
\usepackage{algorithm}
\usepackage{algorithmicx}
\usepackage{algpseudocode}
\usepackage[hidelinks]{hyperref} 
\usepackage{tablefootnote}
\usepackage[misc]{ifsym}
\usepackage{numprint}
\npdecimalsign{.}
\newcounter{rotatenote}

\let\vv\relax
\let\Zbar\relax
\newcommand\paragraphheader[1]{\medskip\noindent \textbf{#1}}
\newcommand\ak[1]{}
\newcommand\fn[1]{}
\newcommand\rg[1]{}
\newcommand{\eat}[1]{} 

\newcommand{\sz}[1]{\lvert#1\rvert}   

\newcommand{\td}[2]{\if*#1\else^{#1}\fi\if*#2\else_{#2}\fi} 

\newcommand\join\Join 

\DeclareSymbolFont{txsymbolsC}{U}{txsyc}{m}{n}
\SetSymbolFont{txsymbolsC}{bold}{U}{txsyc}{bx}{n}
\DeclareFontSubstitution{U}{txsyc}{m}{n}
\DeclareMathSymbol{\ljoin}{\mathrel}{txsymbolsC}{88}
\DeclareMathSymbol{\rjoin}{\mathrel}{txsymbolsC}{89}

\newsavebox\setminusbox
\newlength\setminuslen

\newcolumntype{C}{>{$\displaystyle}c<{$}} 
\newcolumntype{L}{>{$\displaystyle}l<{$}} 
\newcolumntype{R}{>{$\displaystyle}r<{$}} 
\newcolumntype{H}{>{\setbox0=\hbox\bgroup}c<{\egroup}@{}} 

\newcommand{\B}[3]{B\if*#1\else_{#1}\fi(#2,#3)} 
\newcommand{\I}[3]{I\if*#1\else_{#1}\fi(#2,#3)} 

\makeatletter
\def\imod#1{\allowbreak\mkern10mu({\operator@font mod}\,\,#1)}
\makeatother

\newlength\hspaceoflen

\newcommand\vect[1]{{\boldsymbol{#1}}}
\newcommand\va{\vect{a}}
\newcommand\vb{\vect{b}}
\newcommand\vc{\vect{c}}
\newcommand\vd{\vect{d}}
\newcommand\ve{\vect{e}}
\newcommand\vf{\vect{f}}
\newcommand\vg{\vect{g}}
\newcommand\vh{\vect{h}}
\newcommand\vi{\vect{i}}
\newcommand\vj{\vect{j}}
\newcommand\vk{\vect{k}}
\newcommand\vl{\vect{l}}
\newcommand\vm{\vect{m}}
\newcommand\vn{\vect{n}}
\newcommand\vo{\vect{o}}
\newcommand\vp{\vect{p}}
\newcommand\vq{\vect{q}}
\newcommand\vr{\vect{r}}
\newcommand\vs{\vect{s}}
\newcommand\vt{\vect{t}}
\newcommand\vu{\vect{u}}
\newcommand\vv{\vect{v}}
\newcommand\vw{\vect{w}}
\newcommand\vx{\vect{x}}
\newcommand\vy{\vect{y}}
\newcommand\vz{\vect{z}}

\newcommand\mA{\vect{A}}
\newcommand\mB{\vect{B}}
\newcommand\mC{\vect{C}}
\newcommand\mD{\vect{D}}
\newcommand\mE{\vect{E}}
\newcommand\mF{\vect{F}}
\newcommand\mG{\vect{G}}
\newcommand\mH{\vect{H}}
\newcommand\mI{\vect{I}}
\newcommand\mJ{\vect{J}}
\newcommand\mK{\vect{K}}
\newcommand\mL{\vect{L}}
\newcommand\mM{\vect{M}}
\newcommand\mN{\vect{N}}
\newcommand\mO{\vect{O}}
\newcommand\mP{\vect{P}}
\newcommand\mQ{\vect{Q}}
\newcommand\mR{\vect{R}}
\newcommand\mS{\vect{S}}
\newcommand\mT{\vect{T}}
\newcommand\mU{\vect{U}}
\newcommand\mV{\vect{V}}
\newcommand\mW{\vect{W}}
\newcommand\mX{\vect{X}}
\newcommand\mY{\vect{Y}}
\newcommand\mZ{\vect{Z}}

\newcommand\bN{\mathbb{N}} 

\newcommand\bR{\mathbb{R}} 

\DeclareMathAlphabet{\mathcal}{OMS}{cmsy}{m}{n}

\newcommand\cE{\mathcal{E}}

\newcommand\cG{\mathcal{G}}

\newcommand\cK{\mathcal{K}}

\newcommand\cR{\mathcal{R}}

\DeclareMathAlphabet\mathbfcal{OMS}{cmsy}{b}{n}

\accentedsymbol\Abar{{\bar A}}
\accentedsymbol\Bbar{{\bar B}}
\accentedsymbol\Cbar{{\bar C}}
\accentedsymbol\Dbar{{\bar D}}
\accentedsymbol\Ebar{{\bar E}}
\accentedsymbol\Fbar{{\bar F}}
\accentedsymbol\Gbar{{\bar G}}
\accentedsymbol\Hbar{{\bar H}}
\accentedsymbol\Ibar{{\bar I}}
\accentedsymbol\Jbar{{\bar J}}
\accentedsymbol\Kbar{{\bar K}}
\accentedsymbol\Lbar{{\bar L}}
\accentedsymbol\Mbar{{\bar M}}
\accentedsymbol\Nbar{{\bar N}}
\accentedsymbol\Obar{{\bar O}}
\accentedsymbol\Pbar{{\bar P}}
\accentedsymbol\Qbar{{\bar Q}}
\accentedsymbol\Rbar{{\bar R}}
\accentedsymbol\Sbar{{\bar S}}
\accentedsymbol\Tbar{{\bar T}}
\accentedsymbol\Ubar{{\bar U}}
\accentedsymbol\Vbar{{\bar V}}
\accentedsymbol\Wbar{{\bar W}}
\accentedsymbol\Xbar{{\bar X}}
\accentedsymbol\Ybar{{\bar Y}}
\accentedsymbol\Zbar{{\bar Z}}

\accentedsymbol\abar{{\bar a}}
\accentedsymbol\bbar{{\bar b}}
\accentedsymbol\cbar{{\bar c}}
\accentedsymbol\dbar{{\bar d}}
\accentedsymbol\ebar{{\bar e}}
\accentedsymbol\fbar{{\bar f}}
\accentedsymbol\gbar{{\bar g}}
\makeatletter
\@ifundefined{hbar}{}{
        \let\hbar\@undefined
}
\makeatother
\accentedsymbol\hbar{{\bar h}}
\accentedsymbol\ibar{{\bar i}}
\accentedsymbol\jbar{{\bar j}}
\accentedsymbol\kbar{{\bar k}}
\accentedsymbol\lbar{{\bar l}}
\accentedsymbol\mbar{{\bar m}}
\accentedsymbol\nbar{{\bar n}}
\makeatletter
\@ifundefined{obar}{}{
        \let\obar\@undefined
}
\makeatother
\accentedsymbol{\obar}{{\bar o}}
\accentedsymbol\pbar{{\bar p}}
\accentedsymbol\qbar{{\bar q}}
\accentedsymbol\rbar{{\bar r}}
\accentedsymbol\sbar{{\bar s}}
\accentedsymbol\tbar{{\bar t}}
\accentedsymbol\ubar{{\bar u}}
\accentedsymbol\vbar{{\bar v}}
\accentedsymbol\wbar{{\bar w}}
\accentedsymbol\xbar{{\bar x}}
\accentedsymbol\ybar{{\bar y}}
\accentedsymbol\zbar{{\bar z}}

\accentedsymbol\mAhat{{\hat\mA}}
\accentedsymbol\mBhat{{\hat\mB}}
\accentedsymbol\mChat{{\hat\mC}}
\accentedsymbol\mDhat{{\hat\mD}}
\accentedsymbol\mEhat{{\hat\mE}}
\accentedsymbol\mFhat{{\hat\mF}}
\accentedsymbol\mGhat{{\hat\mG}}
\accentedsymbol\mHhat{{\hat\mH}}
\accentedsymbol\mIhat{{\hat\mI}}
\accentedsymbol\mJhat{{\hat\mJ}}
\accentedsymbol\mKhat{{\hat\mK}}
\accentedsymbol\mLhat{{\hat\mL}}
\accentedsymbol\mMhat{{\hat\mM}}
\accentedsymbol\mNhat{{\hat\mN}}
\accentedsymbol\mOhat{{\hat\mO}}
\accentedsymbol\mPhat{{\hat\mP}}
\accentedsymbol\mQhat{{\hat\mQ}}
\accentedsymbol\mRhat{{\hat\mR}}
\accentedsymbol\mShat{{\hat\mS}}
\accentedsymbol\mThat{{\hat\mT}}
\accentedsymbol\mUhat{{\hat\mU}}
\accentedsymbol\mVhat{{\hat\mV}}
\accentedsymbol\mWhat{{\hat\mW}}
\accentedsymbol\mXhat{{\hat\mX}}
\accentedsymbol\mYhat{{\hat\mY}}
\accentedsymbol\mZhat{{\hat\mZ}}

\accentedsymbol\vahat{{\hat\va}}
\accentedsymbol\vbhat{{\hat\vb}}
\accentedsymbol\vchat{{\hat\vc}}
\accentedsymbol\vdhat{{\hat\vd}}
\accentedsymbol\vehat{{\hat\ve}}
\accentedsymbol\vfhat{{\hat\vf}}
\accentedsymbol\vghat{{\hat\vg}}
\accentedsymbol\vhhat{{\hat\vh}}
\accentedsymbol\vihat{{\hat\vi}}
\accentedsymbol\vjhat{{\hat\vj}}
\accentedsymbol\vkhat{{\hat\vk}}
\accentedsymbol\vlhat{{\hat\vl}}
\accentedsymbol\vmhat{{\hat\vm}}
\accentedsymbol\vnhat{{\hat\vn}}
\accentedsymbol\vohat{{\hat\vo}}
\accentedsymbol\vphat{{\hat\vp}}
\accentedsymbol\vqhat{{\hat\vq}}
\accentedsymbol\vrhat{{\hat\vr}}
\accentedsymbol\vshat{{\hat\vs}}
\accentedsymbol\vthat{{\hat\vt}}
\accentedsymbol\vuhat{{\hat\vu}}
\accentedsymbol\vvhat{{\hat\vv}}
\accentedsymbol\vwhat{{\hat\vw}}
\accentedsymbol\vxhat{{\hat\vx}}
\accentedsymbol\vyhat{{\hat\vy}}
\accentedsymbol\vzhat{{\hat\vz}}

\usepackage[outdir=./]{epstopdf}
\usepackage{textcomp}

\begin{document}
\title{Start Small, Think Big:\\On Hyperparameter Optimization\\for Large-Scale Knowledge Graph Embeddings}
\titlerunning{Start Small, Think Big: On HPO for Large-Scale KGEs}
\author{Adrian Kochsiek(\Letter)
\and Fritz Niesel
\and Rainer Gemulla
}
\authorrunning{A. Kochsiek et al.}
\institute{University of Mannheim, Germany \\
\email{\{akochsiek,fniesel,rgemulla\}@uni-mannheim.de}
}
\toctitle{Start Small, Think Big: Hyperparameter Optimization for Large-Scale Knowledge Graph Embeddings}
\tocauthor{Adrian~Kochsiek}
\maketitle              %
\begin{abstract}
  Knowledge graph embedding (KGE) models are an effective and popular approach
  to represent and reason with multi-relational data. Prior studies have shown
  that KGE models are sensitive to hyperparameter settings, however, and that
  suitable choices are dataset-dependent. In this paper, we explore
  hyperparameter optimization (HPO) for very large knowledge graphs, where the
  cost of evaluating individual hyperparameter configurations is excessive.
  Prior studies often avoided this cost by using various heuristics; e.g., by
  training on a subgraph or by using fewer epochs. We systematically discuss and
  evaluate the quality and cost savings of such heuristics and other low-cost
  approximation techniques. Based on our findings, we introduce
  \textsc{GraSH},\footnote{Source code and auxiliary material at
    \url{https://github.com/uma-pi1/GraSH}.} an efficient multi-fidelity HPO
  algorithm for large-scale KGEs that combines both graph and epoch reduction
  techniques and runs in multiple rounds of increasing fidelities. We conducted
  an experimental study and found that \textsc{GraSH} obtains state-of-the-art results on
  large graphs at a low cost (three complete training runs in total).

  \keywords{knowledge graph embedding \and multi-fidelity hyperparameter
    optimization \and low-fidelity approximation}
\end{abstract}
\section{Introduction}
A knowledge graph (KG) is a collection of facts describing relationships between
a set of entities. Each fact can be represented as a (subject, relation,
object)-triple such as (\emph{Rami Malek}, \emph{starsIn}, \emph{Mr.~Robot}).
Knowledge graph embedding (KGE)
models~\cite{bordes2013translating,ConvE,nickel2011three,sun2018rotate,Complex,yang2015embedding}
represent each entity and each relation of the KG with an \emph{embedding},
i.e., a low-dimensional continuous representation. The embeddings are used to
reason about or with the KG; e.g., to predict missing facts in an incomplete
KG~\cite{nickel2015review}, for drug discovery in a biomedical
KG~\cite{mohamed2019drug}, for question answering~\cite{saxena2022sequence,saxena2020improving}, or visual relationship
detection~\cite{baier2017improving}.

Prior studies have shown that embedding quality is highly sensitive to the
hyperparameter choices used when training the KGE
model~\cite{ali2021bringing,ruffinelli2020you}. Moreover, the search space is
large and hyperparameter choices are dataset- and model-dependent.
For example, the best configuration found for one model may perform badly for a different one.
As a consequence, we generally cannot transfer suitable hyperparameter configurations
from one dataset to another or from one KGE model to another. Instead, a separate hyperparameter search is often necessary to achieve high-quality embeddings.

While using an extensive hyperparameter search may be feasible for smaller
datasets---e.g., the study of Ruffinelli et al.~\cite{ruffinelli2020you} uses 200
configurations per dataset and model---, such an approach is generally not
cost-efficient or even infeasible on large-scale KGs, where KGE
training is expensive in terms of runtime, memory consumption, and storage cost.
For example, the Freebase KG consists of $\approx86\,\text{M}$ entities and more than
$300\,\text{M}$ triples.
A single training run of a 512-dimensional ComplEx embedding model on Freebase takes up to $50\,\text{min}$ per epoch utilizing 4~GPUs and requires $\approx164\text{\,GB}$ of memory to store the model.

To reduce these excessive costs, prior studies on large-scale KGE models either
avoid hyperparameter optimization (HPO) altogether or reduce runtime and memory
consumption by employing various heuristics. The former approach leads to
suboptimal quality, whereas the impact in terms of quality and cost of the
heuristics used in the latter approach has not been studied in a principled way.
The perhaps simplest of such heuristics is to evaluate a given hyperparameter
configuration using only a small number of training epochs (e.g.,
\cite{kochsiek2021parallel} uses only 20 epochs for HPO on the Wikidata5M
dataset). Another approach is to use a small subset of the large
KG (e.g., the small FB15k benchmark dataset instead of full Freebase) to obtain
a suitable hyperparameter
configuration~\cite{kochsiek2021parallel,lerer2019pytorch,Zheng2020DGLKETK} or a
set of candidate configurations~\cite{zhang2022efficient}. The general idea
behind these heuristics is to employ \emph{low-fidelity
  approximations} (fewer epochs, smaller graph) to
compare the performance of different hyperparameter configuration during HPO,
before training the final model on \emph{full fidelity} (many epochs, entire graph).

In this paper, we explore how to effectively use a given HPO budget to obtain a
high-quality KGE model. To do so, we first summarize and analyze both cost and
quality of various low-fidelity approximation techniques. We found that there
are substantial differences between techniques and that a combination of
reducing the number of training epochs and the graph size is generally
preferable. To reduce KG size, we propose to use its \emph{$k$-core}
subgraphs~\cite{seidman1983network}; this simple approach worked best throughout our study.

Building upon these results, we present \textsc{GraSH}, an efficient HPO
algorithm for large-scale KGE models. At its heart, \textsc{GraSH} is based on
successive halving~\cite{jamieson2016non}.
It uses multiple fidelities and employs several KGE-specific techniques, most notably, a simple cost model, negative sample
scaling, subgraph validation, and a careful choice of fidelities. We conducted
an extensive experimental study and found that \textsc{GraSH} achieved
state-of-the-art results on large-scale KGs with a low overall search budget
corresponding to only three complete training runs. Moreover, both the use of
multiple reduction techniques simultaneously and of multiple fidelity levels was
key for reaching high quality and low resource consumption.

\section{Preliminaries and Related Work}
\label{sec:background}

A general discussion of KGE models and training is given in~\cite{nickel2015review,wang2017knowledge}.
Here we summarize key points and briefly discuss prior approaches to HPO.

\paragraphheader{Knowledge graph embeddings.} A \emph{knowledge graph}
$\cG=(\cE,\cR,\cK)$ consists of a set $\cE$ of entities, a set $\cR$ of
relations, and a set $\cK \subseteq \cE \times \cR \times \cE$ of triples.
\emph{Knowledge graph embedding} models~\cite{bordes2013translating,ConvE,nickel2011three,sun2018rotate,Complex,yang2015embedding} represent each entity $i \in \cE$ and each relation
$p \in \cR$ with an \emph{embedding} $\ve_i \in \bR^d$ and $\ve_p\in\bR^d$, respectively. They model the
plausibility of each subject-predicate-object triple $(s,p,o)$ via a
model-specific scoring function $f(\ve_s, \ve_p, \ve_o)$, where high scores correspond
to more, low scores to less plausible triples.

\paragraphheader{Training and training cost.} KGE models are
trained~\cite{wang2017knowledge} to provide high scores for the positive triples
in $\cK$ and low scores for negative triples by minimizing a loss such as cross-entropy loss.
Since negatives are typically unavailable, KGE training methods
employ \emph{negative sampling} to generate \emph{pseudo-negative triples},
i.e., triples that are likely but not guaranteed to be actual negatives. The
number $N^-$ of generated pseudo-negatives per positive is an important
hyperparameter influencing both model quality and training cost. In particular,
during each epoch of training a KGE model, all positives and their associated
negatives are scored, i.e., the overall number of per-epoch score computations
is $(\sz{\cK}+1)N^-$. We use this number as a proxy for computational cost
throughout. The size of the KGE model itself scales linearly with the number of
entities and relations, i.e., $O(\sz{\cE}d+\sz{\cR}d)$ if all embeddings are
$d$-dimensional.

\paragraphheader{Evaluation and evaluation cost.} The standard approach to
evaluate KGE model quality for link prediction task is to use the \emph{entity
  ranking} protocol and a filtered metric such as mean reciprocal rank (MRR).
For each $(s,p,o)$-triple in a held-out test set $\cK^{\text{test}}$, this
protocol requires to score all triples of form $(s,p,?)$ and $(?,p,o)$ using all
entities in $\cE$. Overall, $\sz{\cK^{\text{test}}}\sz{\cE}$ scores are computed
so that evaluation cost scales linearly with the number of entities. Since this
cost can be substantial, sampling-based approximations have been used in some
prior studies~\cite{lerer2019pytorch,Zheng2020DGLKETK}. We do not use such
approximations here since they can be misleading in that they do not reflect
model quality faithfully~\cite{kochsiek2021parallel}.

\paragraphheader{Hyperparameters.} The hyperparameter space for KGE models is
discussed in detail in~\cite{ali2021bringing,ruffinelli2020you}. Important hyperparameters
include embedding dimensionalities, training type, number $N^-$ of negatives,
sampling type, loss function, optimizer, learning rate, type and weight of
regularization, and amount of dropout.

\paragraphheader{Full-fidelity HPO.} Recent studies analyzed the impact of
hyperparameters and training techniques for KGE models using full-fidelity
HPO~\cite{ali2021bringing,ruffinelli2020you}. In these studies, the vast
hyperparameter search space was explored using a random search and Bayesian
optimization with more than $200$ full training runs per model and dataset. The
studies focus on smaller benchmark KGs, however; such an approach is excessive
for large-scale knowledge graphs.

\paragraphheader{Low-fidelity HPO.} Current work on large-scale KGE models
circumvented the high cost of full-fidelity HPO by relying on low-fidelity
approximations such as epoch
reduction~\cite{broscheit2020libkge,kochsiek2021parallel} and using smaller
benchmark graphs~\cite{kochsiek2021parallel,lerer2019pytorch,Zheng2020DGLKETK}
in a heuristic fashion. The best performing hyperparameters in low-cost
approximations were directly applied to train a single full-fidelity model. Our
experimental study suggests that such an approach may neither be cost-efficient
nor produce high-quality results.

\paragraphheader{Two-stage HPO.} AutoNE~\cite{tu2019autone} is an HPO approach
for training large-scale network embeddings that optimizes hyperparameters in
two stages. It first approximates hyperparameter performance on subgraphs
created by random walks, a technique that we will explore in Sec.~\ref{sec:fidelity_techniques}.
Subsequently, AutoNE transfers these results to the full graph using a meta
learner. In the context of KGs, this approach was outperformed by
KGTuner~\cite{zhang2022efficient},\footnote{KGTuner was proposed in parallel to
  this work.} which uses a multi-start random walk (fixed to $20\%$ of the
entities) in the first stage and evaluates the top-performing configurations
(fixed to $10$) at full fidelity in the second stage.
Such fixed heuristics often limit flexibility in terms of budget allocation and lead to an expensive second stage on large KGs.
In contrast, \textsc{GraSH} makes use of
multiple fidelity levels, carefully constructs and evaluates low-fidelity
approximations, and adheres to a prespecified overall search budget. These
properties are key for large KGs; see Sec.~\ref{sec:model_quality}
for an experimental comparison with KGTuner.

\section{Successive Halving for Knowledge Graphs (\textsc{GraSH})}
\label{sec:multifidelity_hpo}
\label{sec:our_approach}

\textsc{GraSH} is a multi-fidelity HPO algorithm for KGE models based on
successive halving~\cite{jamieson2016non}. As successive halving, \textsc{GraSH}
proceeds in multiple \emph{rounds} of increasing fidelity; only the best
configurations from each round are transferred to the next round. In contrast to
the HPO techniques discussed before, this approach allows to discard unpromising
configurations at very low cost. \textsc{GraSH} differs from successive halving
mainly in its parameterization and its use of KG-specific reduction and
validation techniques.

\paragraphheader{Parameterization.} \textsc{GraSH} is summarized as
Alg.~\ref{alg:successive_halving}. Given knowledge graph~$\cG$, \textsc{GraSH}
outputs a single optimized hyperparameter configuration. \textsc{GraSH} is
parameterized as described in Alg.~\ref{alg:successive_halving}; default
parameter values are given in parentheses if applicable. The most important
parameters are the maximal number $E$ of epochs and the overall search budget
$B$. The search budget $B$ is relative to the cost of a full training run, which
in turn is determined by $E$. The default choice $B=3$, for example, corresponds
to an overall search cost of three full training runs. We chose this
parameterization because it is independent of utilized hardware and both
intuitive and well-controllable. The reduction factor~$\eta$ controls the number of
configurations (starts at $n$, decreases by factor of $\eta$ per round) and fidelity
(increases by factor of $\eta$) of each round. Note that \textsc{GraSH} does not
train at full fidelity, i.e., its final configuration still needs to be trained
on the full KG (not part of budget $B$). Finally, \textsc{GraSH} is
parameterized by a \emph{variant}~$v$. This parameter controls which reduction
technique to use (only epoch, only graph, or combined).

\begin{algorithm}[t]
  \caption[GraSH]{\textsc{GraSH}: Successive Halving for Knowledge Graph
    Embeddings}%
\begin{algorithmic}
\Require \\
KG $\cG=(\cE,\cR,\cK)$,
max.~epochs $E$,
search budget $B$ (=3),
num.~configurations~$n$ (=64),
reduction factor $\eta$ (=4),
variant $v \in \{$\emph{epoch}, \emph{graph}, \emph{combined}$\}$ (=\emph{combined})
\Ensure Hyperparameter configuration
\end{algorithmic}
\begin{algorithmic}[1]
\State $s \gets \lceil \log_\eta(n)\rceil$ \Comment{Number of rounds}
\State $R \gets B/s$ \Comment{Per-round budget}
\State $\Lambda_1 \gets \{\lambda_1, ..., \lambda_{n}\}$ \Comment{Generate $n$ hyperparameter configurations}
\For{$i \in \{1, ..., s\}$} \Comment $i$-th round

  \State $f_i\gets R/|\Lambda_i|$ \textbf{if} $v\neq\textit{combined}$ \textbf{else} $R/\sqrt{|\Lambda_i|}$\Comment{Target fidelity}\label{alg:reduct_per_trial}
\State $E_i \gets f_i E$ \textbf{if} $v\ne$ \emph{graph} \textbf{else} $E$ \Comment{Epochs in round $i$}
\State  $\cG_i \gets$ reduced KG with $f_i|\cK|$ triples \textbf{if} $v\ne$ \emph{epoch} \textbf{else} $\cG$ \Comment{Graph in round $i$}
\State $\cG_i^{\text{train}}, \cG_i^{\text{valid}} \gets$ random train-valid split of $\cG_i$

\State $V_i \gets $ train each $\lambda\in \Lambda_i$ on $\cG_i^{\text{train}}$ for $E_i$ epochs and validate using $\cG_i^{\text{valid}}$ \label{alg:train}
\State $\Lambda_{i+1} \gets $ best $\lceil |\Lambda_i|/\eta \rceil$ configurations from $\Lambda_i$ according to $V_i$ \label{alg:select}
\EndFor
\State \textbf{return} $\Lambda_{s+1}$ \label{alg:return} \Comment{Only single configuration left}

\end{algorithmic}
\label{alg:successive_halving}
\end{algorithm}

\paragraphheader{Algorithm overview.} Like successive halving, \textsc{GraSH}
proceeds in rounds. Each round has approximately the same overall budget, but
differs in the number of configurations and fidelity. For example, using the default
settings of $B=3$, $n=64$ and $\eta=4$, \textsc{GraSH} uses three rounds with 64,
16, and 4 configurations and a fidelity of 1/64, 1/16, 1/4, respectively. The
hyperparameter configurations in the first round are sampled randomly from the
hyperparameter space.
Depending on the variant being used, \textsc{GraSH} reduces the number of epochs, the graph size, or both to reach the desired fidelity.
If no reduced graph corresponds to the fidelity, the next smaller one is used.
After validating each configuration (see below), the best performing
$1/\eta$-th of the configurations is passed on to the next round. This process is
repeated until only one configuration remains.

\paragraphheader{Validation on subgraphs.} Care must be taken when validating a
KGE model trained on a subgraph, e.g., $\cG_i=(\cE_i,\cR_i,\cK_i)$ in round $i$.
Since $\cG_i$ typically contains a reduced set of entities $\cE_i\subseteq \cE$,
a full validation set for $\cG$ cannot be used. This is because no embedding is
learned for the ``unseen'' entities in $\cE\setminus \cE_i$, so that we cannot score any
triples containing these entities (as required by the entity ranking protocol).
To avoid this problem, we explicitly create new train and valid splits
$\cG_i^{\text{train}}$ and $\cG_i^{\text{valid}}$ in round $i$.
Here, $\cK_i^{\text{valid}}$ is sampled randomly from $\cK_i$ and $\cK_i^{\text{train}}=\cK_i\setminus \cK_i^{\text{valid}}$.
Although this approach is very simple, it worked well in our study. An alternative is the
construction of ``hard'' validation sets as in~\cite{toutanova2015observed}. We
leave the exploration of such techniques to future work.

\paragraphheader{Negative sample scaling.} Recall that the number $N^-$ of
negative samples is an important hyperparameter for KGE model training.
Generally (and assuming without-replacement sampling), each entity is sampled as
a negative with probability $N^-/\sz{\cE}$. When we use a subgraph $\cG_i$ as in
\textsc{GraSH}, this probability increases to $N^-/\sz{\cE_i}$, i.e., each
entity is more likely to act as a negative sample due to the reduction of the
number of entities. To correctly assess hyperparameter configurations in such
cases, \textsc{GraSH} scales the number of negative examples and uses
$N^-_i=\frac{\sz{\cE_{i}}}{\sz{\cE}} N^-$ in round $i$. This choice preserves
the probability of sampling each entity as a negative and provides additional
cost savings since the total number of scored triples is further reduced in
low-fidelity experiments.

\paragraphheader{Cost model and budget allocation.} To distribute the search
budget $B$ over the rounds, we make use of a simple cost model to estimate the
relative runtime of low-fidelity approximations. This cost model drives the
choice of $f_i$ in Alg.~\ref{alg:successive_halving}. In particular, we assume
that training cost is linear in both the number of epochs ($E_i$) and the number
of triples ($\sz{\cK_i}$). For example, this implies that training five
configurations for one epoch has the same cost as training one configuration for
five epochs. Likewise, training five configurations with 20\% of the triples has
the same cost as training one configuration on the whole KG. Using this
assumption, the relative cost of evaluating a single hyperparameter
configuration in round $i$ is given by $\frac{E_i}{E}
\frac{\sz{\cK_i}}{\sz{\cK}}$. More elaborate cost models are conceivable, but
this simple approach already worked well in our experimental study. Note, for
example, that our simple cost model neglects negative sample scaling and thus
tends to overestimate (but avoids underestimation) of training cost.

\section{Low-fidelity Approximation Techniques}
\label{sec:fidelity_techniques}

In this section, we summarize and discuss various low-fidelity approximation
techniques. As discussed previously, the two most common types are \emph{graph
  reduction} (i.e. training on a reduced graph) and \emph{epoch reduction}
(i.e., training for fewer epochs). Note that although graph reduction is related to
dataset reduction techniques used in other machine learning domains, it
represents a major challenge since the relationships between entities need to be
taken into account.

Generally, good low-fidelity approximations satisfy the following criteria:
\begin{enumerate}
\item \textbf{Low cost.} Computational and memory costs for model training
  (including model initialization) and evaluation should be low. Recall that
  computational costs are mainly determined by the number of triples, whereas
  memory and evaluation cost are determined by the number of entities. Ideally,
  both quantities are reduced.
\item \textbf{High transferability.} Low-fidelity approximations should transfer to the full KG in that they provide useful information. E.g., rankings of low-fidelity approximations should match or correlate with the rankings at full-fidelity.
\item \textbf{Flexibility.} It should be possible to flexibly trade-off
  computational cost and transferability.
\end{enumerate}
All three points are essential for cost-effective and practical multi-fidelity
HPO.

In the following, we present the graph reduction approaches \emph{triple
  sampling}, \emph{multi-start random walk}, and \emph{$k$-core decomposition},
as well as epoch reduction. A high-level comparison of these approaches
w.r.t.~the above desiderata is provided in Tab.~\ref{tab:fidelity_options}.
The assessment given in the table is based on our experimental results (Sec.~\ref{sec:transferability_exp}).

\begin{table}[t]
	\centering
	\caption{Comparison of low-fidelity approximation techniques.}
	\begin{tabular}{lcccc}
		\toprule
		\multicolumn{1}{l}{\begin{tabular}[c]{@{}l@{}}Technique\end{tabular}} &
		\multicolumn{1}{c}{\begin{tabular}[c]{@{}c@{}}Low\\Cost\end{tabular}} &
		\multicolumn{1}{c}{\begin{tabular}[c]{@{}c@{}}High\\Transferability\end{tabular}} &
		\multicolumn{1}{c}{\begin{tabular}[c]{@{}c@{}}Flexibility\\ \end{tabular}} \\
		\midrule
		Triple sampling & $\circ$ & - & +\\
		Random walk & $\circ$ & $\circ$ & +\\
		$k$-core decomposition & + & + & $\circ$\\
		\midrule
		Epoch reduction  & - & $\circ$ & +\\
		\bottomrule
	\end{tabular}
	\label{tab:fidelity_options}
\end{table}

\subsection{Graph Reduction} \label{sec:graph_reduction}

\begin{figure}[t]
	\centering
	\begin{subfigure}[t]{0.32\textwidth}
		\centering
		\includegraphics[width=0.7\textwidth]{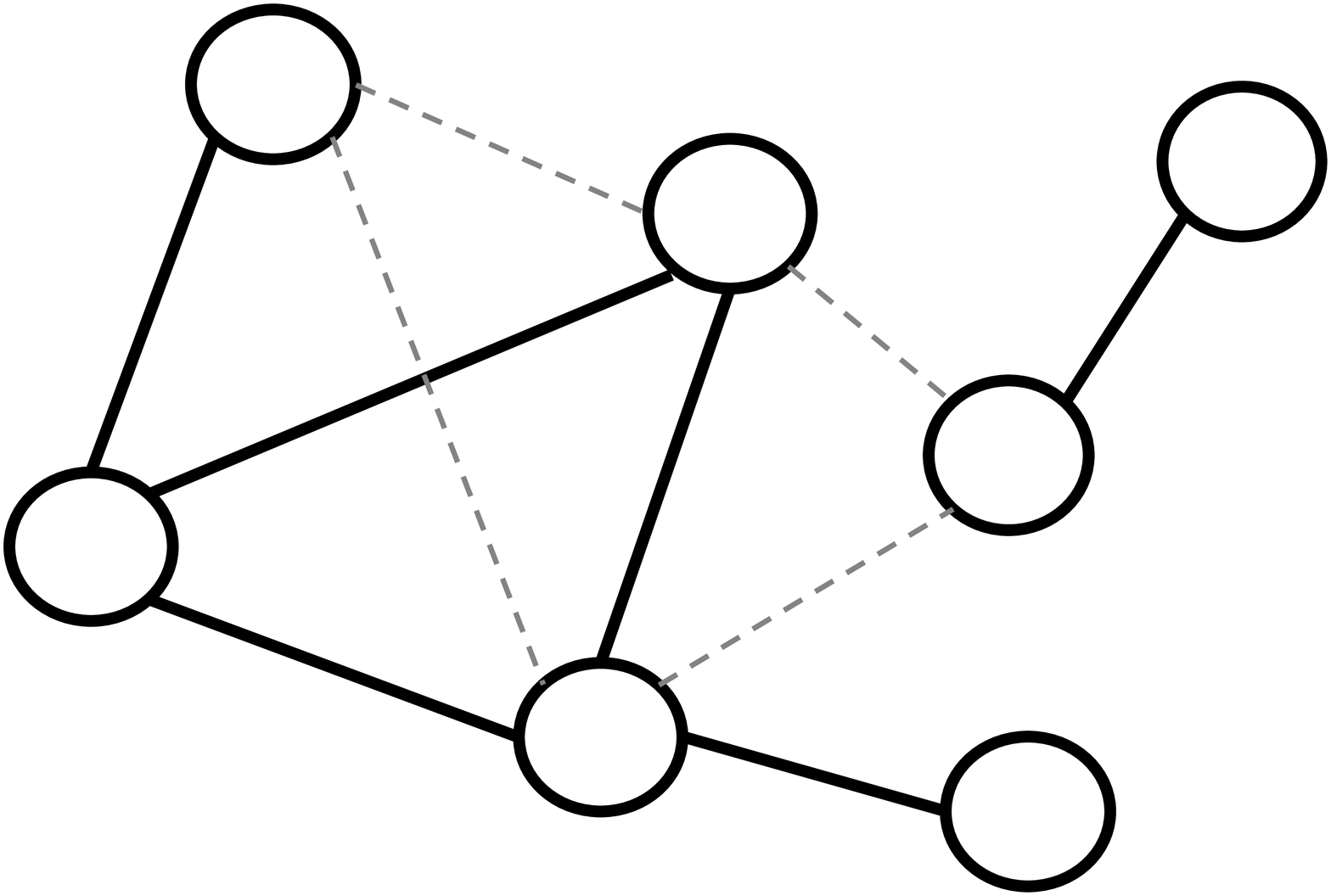}
		\caption{Triple sampling \\ (60\%).}
		\label{fig:triple_sampling}
	\end{subfigure}
		\begin{subfigure}[t]{0.32\textwidth}
		\centering
		\includegraphics[width=0.7\textwidth]{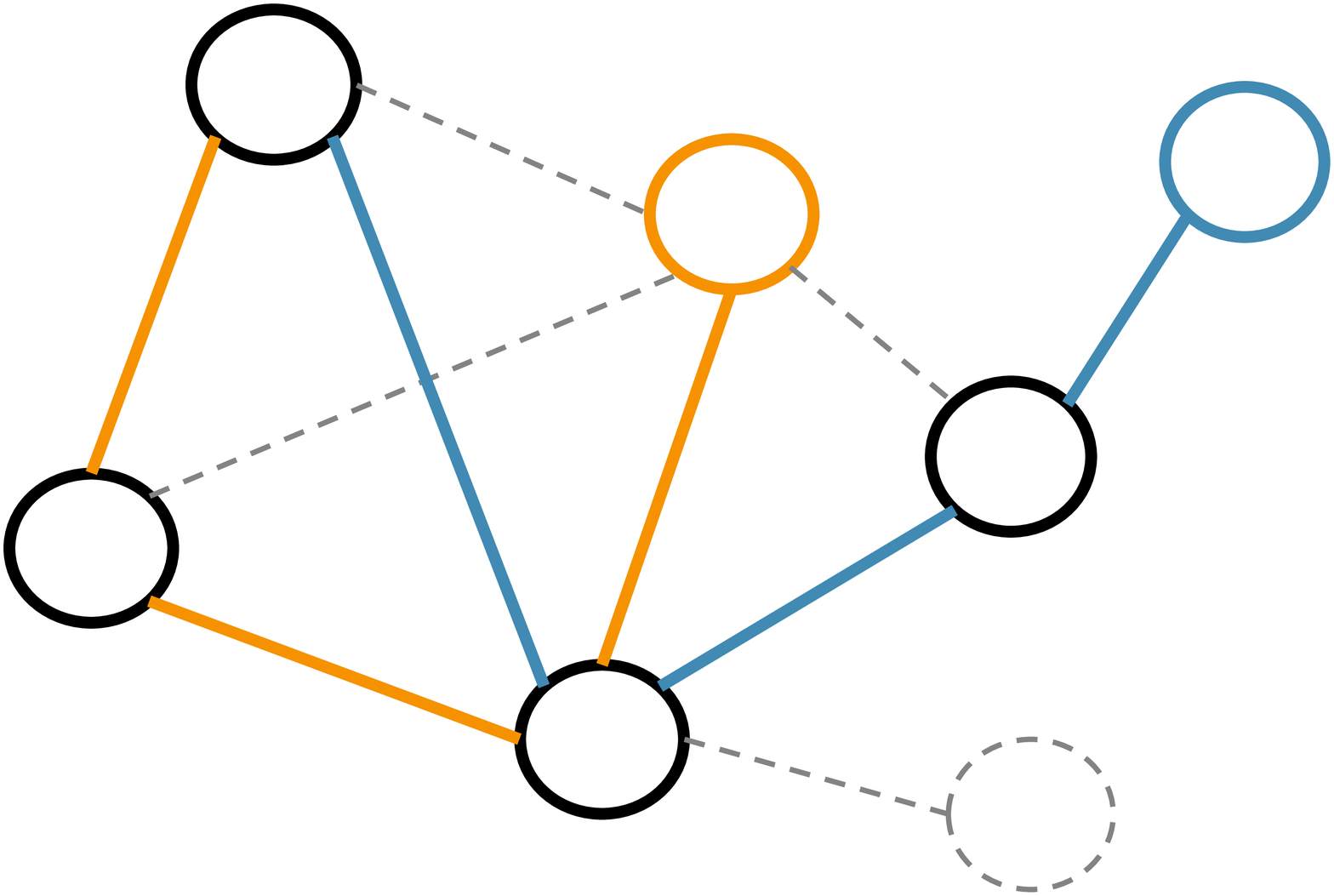}
		\caption{Random walk \\ ($s=2$, $l=3$).}
		\label{fig:random_walk}
	\end{subfigure}
	\begin{subfigure}[t]{0.32\textwidth}
		\centering
		\includegraphics[width=0.7\textwidth]{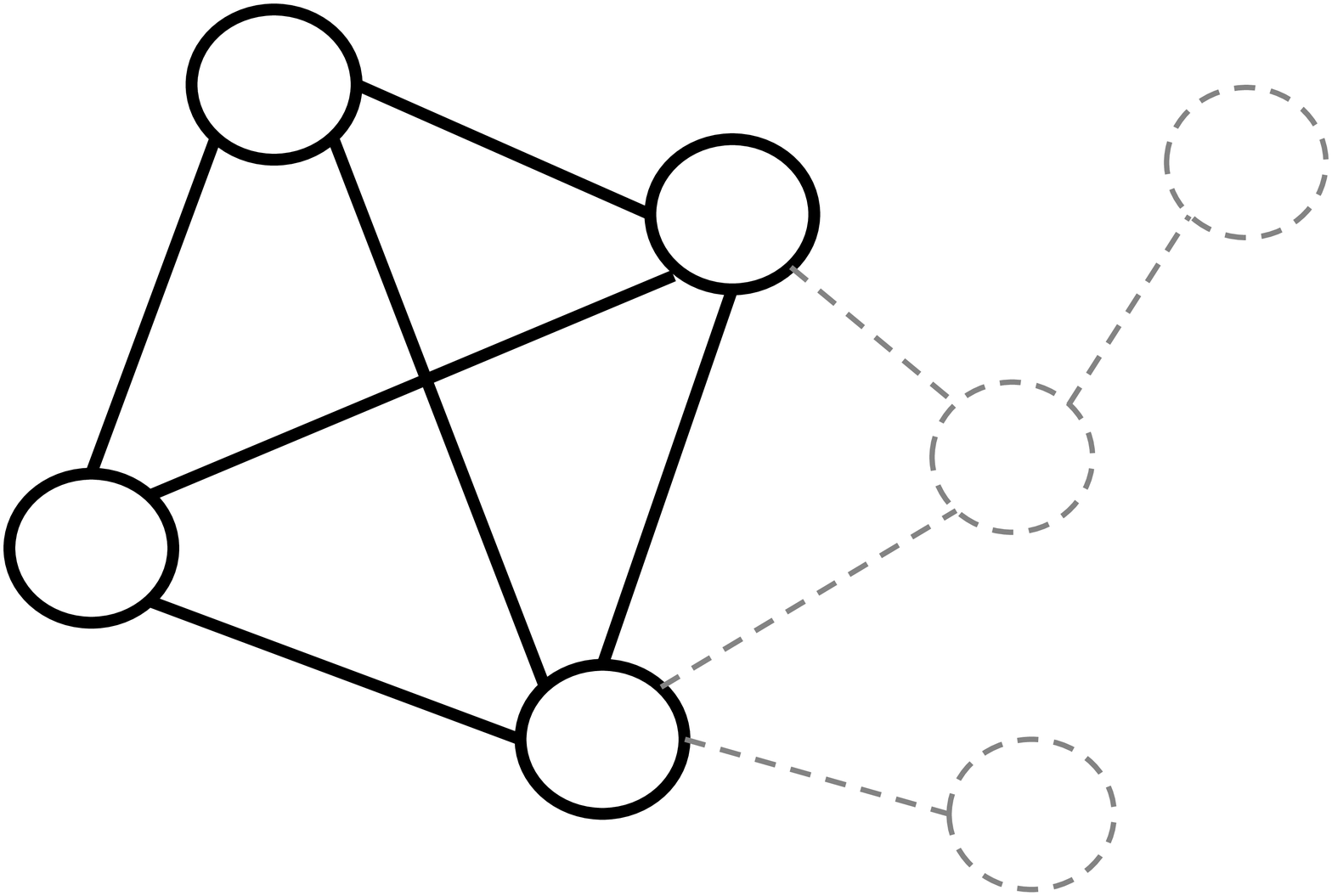}
		\caption{$k$-core decomposition \\ ($k=3$).}
		\label{fig:core}
	\end{subfigure}
	\caption{Schematic illustration of selected graph reduction techniques. All reduced graphs contain 6 of the 10 original triples but a varying number of entities.}
	\label{fig:subgraphs}
\end{figure}

Graph reduction techniques produce a reduced KG $\cG_i=(\cE_i,\cR_i,\cK_i)$ from
the full KG $\cG=(\cE,\cR,\cK)$. This is commonly done by first determining the
reduced set $\cK_i$ of triples and subsequently retaining only those entities
(in $\cE_i$) and relations (in $\cR_i$) that occur in $\cK_i$.\footnote{All
  other entities/relations do not occur in the reduced training data so that we
  cannot learn useful embeddings for them.}
A reduction in triples thus may lead to a reduction in the number of entities and relations as well.
This consequently results in further savings in computational cost, evaluation cost, and memory consumption.
The graph reduction techniques discussed here are illustrated in Fig.~\ref{fig:subgraphs}.

\paragraphheader{Triple sampling (Fig.~\ref{fig:triple_sampling}).} The perhaps
simplest approach to reduce graph size is to sample triples randomly from the
graph. As shown in Fig.~\ref{fig:triple_sampling}, many entities with sparse
interconnections can remain in the resulting subgraphs (e.g., the two entities
at the top right) so that $\cE_i$ tends to be large. The cost in terms of model
size and evaluation time is consequently only slightly reduced. We also
observed (see Sec.~\ref{sec:transferability_exp}) that triple sampling leads to
low transferability, most likely due to this sparsity. Triple sampling does
offer very good flexibility, however, since triple sets of any size can be
constructed easily.

\paragraphheader{Random walk (Fig.~\ref{fig:random_walk}).} In multi-start
random walk, which is used in AutoNE~\cite{tu2019autone}, a set of $s$ random
entities is samples from $\cE$. A random walk of length~$l$ is started from each
of these entities and the resulting triples form $\cK_i$. Empirically, many
entities may ultimately remain so that the reduction of memory consumption and
evaluation cost is limited. Although the resulting subgraph tends to be better
connected than the ones obtained by triple sampling, transferability is still
low and close to triple sampling (again, see
Sec.~\ref{sec:transferability_exp}). As triple sampling, the approach is very
flexible though. KGTuner~\cite{zhang2022efficient} improves on the basic
random walk considered here by using biased starts and adding all connections
between the retained entities (even if they do not occur in a walk). The $k$-core
decomposition, which we discuss next, offers a more direct approach to obtain
such a highly-connected graph.

\paragraphheader{$k$-core decomposition (Fig.~\ref{fig:core}).} The $k$-core
decomposition~\cite{seidman1983network} allows for the construction of subgraphs
with increasing cohesion. The $k$-core subgraph of $\cK$, where $k\in\bN$ is a
parameter, is defined as the largest induced subgraph in which every retained
entity (i.e., $\cE_i$) occurs in at least $k$ retained triples (i.e., in
$\cK_i$). The computation of $k$-cores is cheap and supported by common graph
libraries.
Generally, $k$-cores contain only a small number of entities because long-tail entities with infrequent connections
are removed. Moreover, they are highly interconnected by construction. As a
consequence, we found that computational cost and memory consumption is low and
transferability high. The approach is less flexible than the other graph
reduction techniques, as the choice of $k$ and the graph structure
determines the resulting fidelity. One may interpolate between $k$-cores for
improved flexibility but we do not explore this approach in this work.

\subsection{Epoch Reduction} \label{sec:epoch_reduction}

Epoch reduction is the most common form of fidelity control used in
HPO~\cite{baker2017accelerating,wang2021rank}. As the set $\cE$ of entities does
not change with varying fidelity, memory and evaluation cost are very large even
when using low-fidelity approximations. We observed good transferability as long
as the number of epochs is not too small (Sec.~\ref{sec:transferability_exp});
otherwise, transferability is often considerably worse than graph reduction
techniques. This limits flexibility: Especially on large-scale graphs, the
overall training budget often consists of only a small number of epochs in the
first place (e.g., $10$ as in~\cite{kochsiek2021parallel,Zheng2020DGLKETK}).
Note that the available budget in low-fidelity approximations can be smaller
than the cost of one complete epoch (when $f_i<1/E$ in
Alg.~\ref{alg:successive_halving}). Although partial epochs can be used easily,
epoch reduction then corresponds to a form of triple sampling (with the additional
disadvantage of not reducing the set of entities).

\subsection{Summary}
\label{sec:low_fidelity_discussion}

In summary, as long the desired fidelity is sufficiently high, epoch reduction
offers high-quality approximations and high flexibility. It does not improve
memory consumption and evaluation cost, however, and it leads to high cost and
low quality on large-scale graphs with limited budget.
Graph reduction approaches, on the other hand, reduce the number of entities and
hence memory consumption and evaluation cost. Compared to triple sampling and
random walks, the $k$-core decomposition has the highest transferability and lowest
cost. In \textsc{GraSH}, we use a combination of epoch reduction and $k$-core
decomposition by default to avoid training for partial epochs and the use of very small subgraphs with low-fidelity.

\section{Experimental Study}
We conducted an experimental study to investigate (i) to what extent
hyperparameter rankings obtained with low-fidelity approximations correlate with
the ones obtained at full fidelity (Sec.~\ref{sec:transferability_exp}); (ii)
the performance of \textsc{GraSH} in terms of quality
(Sec.~\ref{sec:model_quality}), resource consumption
(Sec.~\ref{sec:runtime_comparison_exp}) and robustness
(Sec.~\ref{sec:robustness}). In summary, we found that:

\begin{enumerate}
\item \textsc{GraSH} was cost-effective and produced high-quality hyperparameter
  configurations. It reached state-of-the-art results on a large-scale graph
  with a small overall search budget of three complete training runs
  (Sec.~\ref{sec:model_quality}).
\item Using multiple reduction techniques was beneficial. In particular, a
  combination of graph- and epoch-reduction performed best
  (Sec.~\ref{sec:transferability_exp} and~\ref{sec:model_quality}).
\item Low-fidelity approximations correlated best to full fidelity for graph
  reduction using the $k$-core decomposition and, as long as the budget was
  sufficiently large, second-best for epoch reduction
  (Sec.~\ref{sec:transferability_exp}).
\item Graph reduction was more effective than epoch reduction in terms of
  reducing computational and memory cost. Evaluation using small subgraphs had
  low memory consumption and short runtimes
  (Sec.~\ref{sec:runtime_comparison_exp}).
\item Using multiple rounds with increasing fidelity levels was
  beneficial (Sec.~\ref{sec:robustness}).
\item GraSH was robust to changes in budget allocation across rounds (Sec.~\ref{sec:robustness}).
\end{enumerate}

{
    \setlength{\tabcolsep}{2.5pt}
\begin{table}[t]
	\caption{\label{tab:dataset_statistics}Dataset statistics.}
	\centering
	\begin{tabular}{llrrrrr} \toprule
        \multicolumn{1}{l}{Scale} &
        \multicolumn{1}{l}{Dataset} &
		\multicolumn{1}{r}{Entities} &
        \multicolumn{1}{r}{Relations} &
        \multicolumn{1}{r}{$|$Train$|$} &
        \multicolumn{1}{r}{$|$Valid$|$} &
        \multicolumn{1}{r}{$|$Test$|$} \\
        \midrule
        Small & Yago3-10   & \numprint{123182}                      & \numprint{37}                            & \numprint{1079040}     & \numprint{5000}      & \numprint{5000}            \\ %
        Medium & Wikidata5M & \numprint{4594485}                    & \numprint{822}                           & \numprint{21343681}    & \numprint{5357}      & \numprint{5321}           \\ %
        Large & Freebase   & \numprint{86054151}                   & \numprint{14824}                        & \numprint{304727650}   & \numprint{1000} & \numprint{10000} \\ %
        \bottomrule
	\end{tabular}
\end{table}
}

\subsection{Experimental Setup}
Source code, search configurations, resulting hyperparameters, and an online
appendix can be found at \url{https://github.com/uma-pi1/GraSH}.

\paragraphheader{Datasets.} We used common KG benchmark datasets of
varying sizes with a focus on larger datasets; see
Tab.~\ref{tab:dataset_statistics}. \emph{Yago3\nobreakdash-10}~\cite{ConvE} is a
subset of Yago~3 containing only entities that occur at least ten times in
the complete graph. \emph{Wikidata5M}~\cite{wang2019kepler} is a large-scale
benchmark and the induced graph of the five million most-frequent entities of
Wikidata. The largest dataset is \emph{Freebase} as used
in~\cite{kochsiek2021parallel,Zheng2020DGLKETK}. For all datasets except
Freebase, we use the validation and test sets that accompany the datasets to
evaluate the final model. For Freebase, we used the sub-sampled validation
(\numprint{1000} triples) and test sets (\numprint{10000} triples)
from~\cite{kochsiek2021parallel}.\footnote{The original test set contains
  $\approx$$17\text{\,M}$ triples, which leads to excessive evaluation costs. For the
  purpose of MRR computation, a much smaller test set is sufficient.}

\paragraphheader{Hardware.} All runtime, GPU memory, and model size measurements
were taken on the same machine (40 Intel Xeon E5-2640 v4 CPUs @ 2.4GHz; 4 NVIDIA
GeForce RTX 2080 Ti GPUs).

\paragraphheader{Implementation and models.}
\textsc{GraSH} uses \textsc{DistKGE}~\cite{kochsiek2021parallel} for parallel training of large-scale graphs and HpBandSter~\cite{falkner2018bohb} for the implementation of SH.
We considered the models ComplEx~\cite{Complex}, RotatE~\cite{sun2018rotate} and
TransE~\cite{bordes2013translating}. ComplEx and RotatE are among the currently
best-performing KGE
models~\cite{ali2021bringing,lacroix2018canonical,ruffinelli2020you,sun2018rotate}
and represent semantic matching and translational distance models, respectively.
All three models are commonly used for large-scale
KGEs~\cite{kochsiek2021parallel,lerer2019pytorch,Zheng2020DGLKETK}.

\paragraphheader{Hyperparameters.} We used the hyperparameter search space
of~\cite{kochsiek2021parallel}. The search space consists of nine continuous and
two categorical hyperparameters. The upper bound on the number of negative
samples for ComplEx is \numprint{10000} and for RotatE and TransE
\numprint{1000} (since these models are more memory-hungry). We set the maximum
training epochs on Yago3\nobreakdash-10 to $400$, on Wikidata5M to $64$, and on
Freebase to $10$.

\paragraphheader{Methodology.} For the \textsc{GraSH} search, we used the
default settings ($B=3$, $\eta=4$, $n=64$). Apart from the upper bound of
negatives, we used the same $64$ initial hyperparameter settings for all models
and datasets to allow for a fair comparison. For graph reduction, we used
$k$-core decomposition unless mentioned otherwise. Subgraph validation sets generated by \textsc{GraSH} consisted of \numprint{5000} triples.
The resulting best configurations are published along with our \href{https://github.com/uma-pi1/GraSH}{online appendix}.

\begin{figure}[t]
    \centering

    \begin{subfigure}[t]{0.47\textwidth}
        \centering
        \includegraphics[width=0.835\textwidth]{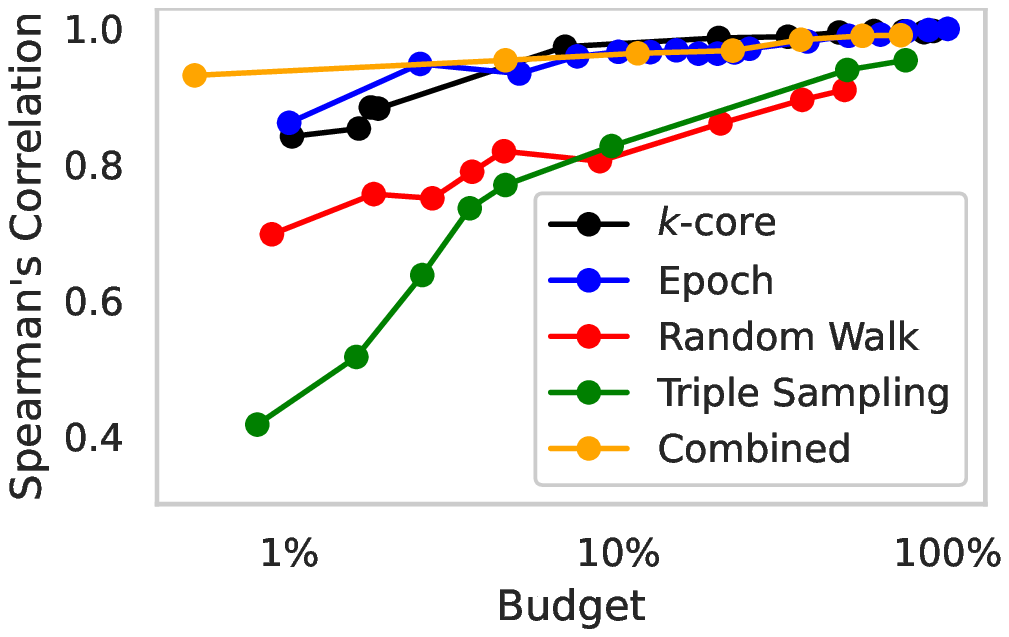}
        \caption{Yago3-10 (max.~$40$ epochs).}
        \label{fig:buget_vs_correlation_yago}
    \end{subfigure}
    \begin{subfigure}[t]{0.51\textwidth}
        \centering
        \includegraphics[width=0.769\textwidth]{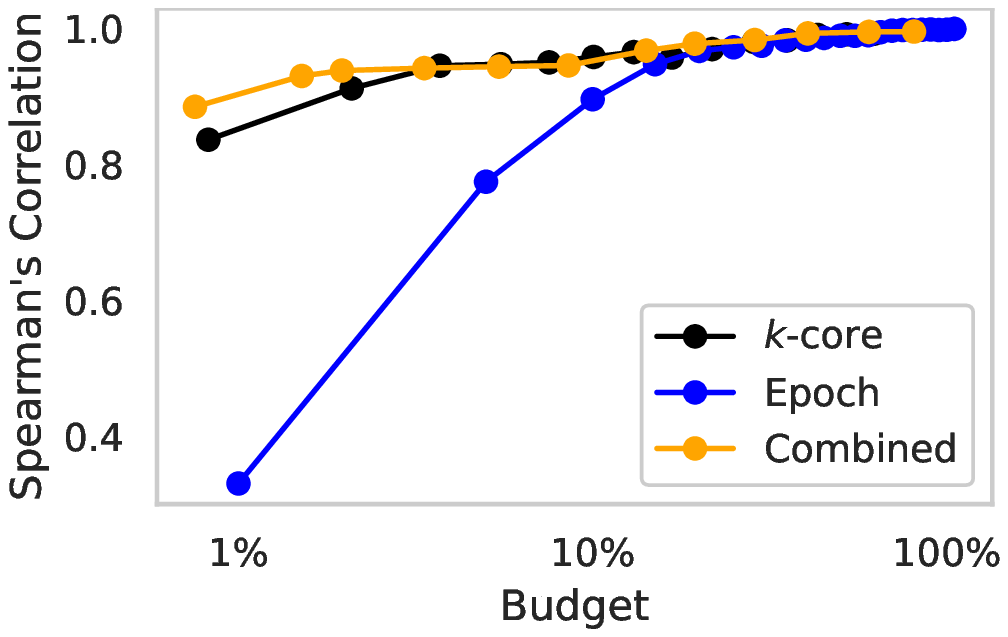}
        \caption{Wikidata5M (max.~$20$ epochs).}
        \label{fig:budget_vs_correlation_wikidata}
    \end{subfigure}
    \caption{Comparison of low-fidelity approximations techniques.
      Shows Spearman's rank correlation between low-fidelity approximations and a full-fidelity baseline. Budget (log-scale) corresponds to the relative amount of epochs and/or triples.
    }
    \label{fig:transferability}
\end{figure}

\paragraphheader{Metrics.} We used the common filtered MRR metric to evaluate
KGE model quality on the link prediction task as described in
Sec.~\ref{sec:background}.
Results for Hits@$k$ are given in our
\href{https://github.com/uma-pi1/GraSH}{online appendix}.
\subsection{Comparison of Low-Fidelity Approximation Techniques (Fig.~\ref{fig:transferability})}
\label{sec:transferability_exp}

In our first experiment, we studied and compared the transferability of
low-fidelity approximations to full-fidelity results. To do so, we first ran a
full-fidelity hyperparameter search consisting of 30 pseudo-randomly generated
trials. We then trained and evaluated the same 30 trials using the approximation
techniques described in Sec.~\ref{sec:fidelity_techniques} at various budgets.
To keep computational cost feasible, this experiment was only performed on the
two smaller datasets and with a smaller number of epochs.

Since the validation sets used with graph reductions differ from the one used at
full fidelity (see Sec.~\ref{sec:multifidelity_hpo}), we compared the ranking of
hyperparameter configurations instead of their MRR metrics. In particular, we
used Spearman's rank correlation coefficient~\cite{zwillinger1999crc} between
the low-fidelity and the high-fidelity results. A higher value corresponds to a
better correlation.

Our results on Yago3\nobreakdash-10 are visualized in
Fig.~\ref{fig:buget_vs_correlation_yago}. We found high transferability for the
$k$-core decomposition and epoch reduction. Graph reduction based on triple
sampling and random walks led to clearly inferior results and was not further
considered. A combination of $k$-core subgraphs and reduced epochs (each
contributing $50$\% to the savings) further improved low-budget results.

To investigate the behavior on a larger graph, we evaluated the three best
techniques on Wikidata5M, see Fig.~\ref{fig:budget_vs_correlation_wikidata}.
Recall that due to the high cost, a small number of epochs is often used for
training on large KGs. This has a detrimental effect on the transferability of
epoch reduction, as partial epochs need to be used for low-fidelity
approximations (see Sec.~\ref{sec:epoch_reduction}). In particular, there is a
considerable drop in transferability for epoch reduction below the 10\% budget. This
drop in performance is neither visible for the $k$-core approximations nor for
the combined approach.

Note that even for the best low-fidelity approximation, the rank correlation
increased with budget. This suggests that using multiple fidelities (as in
\textsc{GraSH}) instead of a single fidelity is beneficial. In our study, this
was indeed the case (see Sec.~\ref{sec:robustness}).

\subsection{Final Model Quality (Tab.~\ref{tab:quality})}
\label{sec:model_quality}

In our next experiment, we analyzed the performance of \textsc{GraSH} in terms
of the quality of its selected hyperparameter configuration.
Tab.~\ref{tab:mvsh_quality} shows the test-data performance of this resulting
configuration trained at full fidelity. We report results for different
datasets, different reduction techniques, different KGE models, and different
model dimensionalities.

\paragraphheader{Results (Tab.~\ref{tab:mvsh_quality}).} The combined variant of
\textsc{GraSH} offered best or close to best results across all datasets and
models. In comparison to the other variants, it avoided the drawbacks of
training partial epochs (e.g., epoch reduction on Freebase) as well as using subgraphs that are too
small (e.g., graph reduction on Yago3\nobreakdash-10). %
\paragraphheader{Comparison to prior results (Tab.~\ref{tab:prior_results}).} We
compared the results obtained by \textsc{GraSH} to the best published prior
results known to us, see Tab.~\ref{tab:prior_results}. Note that prior models
were often trained at substantially higher cost. For example, on Wikidata5M,
\textsc{GraSH} used an overall budget of $4\cdot64=256$ epochs for HPO and training, whereas some prior methods used \numprint{1000}
epochs for a single training run. Likewise, dimensionalities of up to
\numprint{1000} were sometimes used. For a slightly more informative comparison,
we performed a \textsc{GraSH} search with an increased dimensionality of $512$,
but kept the low search and training budgets. Even with this low budget, we
found that on small to midsize graphs, \textsc{GraSH} performed either similarly
(ComplEx, Yago3-10 \& Wikidata5M) and sometimes slightly worse (RotatE,
Wikidata5M) than the best prior results. On the large-scale Freebase KG, where
low-fidelity hyperparameter search is a necessity, \textsc{GraSH} outperformed
state-of-the-art results by a large margin.

    {
    \setlength{\tabcolsep}{0.9pt}
    \begin{table}[t]
        \caption{Model quality in terms of MRR. State-of-the-art results underlined.
        Best reduction variant in bold. Note that best prior results often use a
        considerably larger budget and/or model dimensionality.}
        \begin{subtable}[t]{.71\linewidth}
            \caption{\textsc{GraSH} with default settings ($B=3$, $n=64$, $\eta=4$).}
            \label{tab:mvsh_quality}
            \begin{tabular}{lclrrr@{$\;$}|r}
                \toprule
                & %
                \multicolumn{1}{c}{\begin{tabular}[c]{@{}c@{}}Dataset\end{tabular}} &
                \multicolumn{1}{c}{\begin{tabular}[c]{@{}c@{}}Variant $\to$\\ Model $\downarrow$\end{tabular}} &
                \multicolumn{1}{c}{\begin{tabular}[c]{@{}c@{}}Epoch\\Dim 128\end{tabular}} &
                \multicolumn{1}{c}{\begin{tabular}[c]{@{}c@{}}Graph\\Dim 128\end{tabular}} &
                \multicolumn{1}{c}{\begin{tabular}[c]{@{}c@{}}Comb.\\Dim 128\end{tabular}} &
                \multicolumn{1}{c}{\begin{tabular}[c]{@{}c@{}}Comb.\\Dim 512\end{tabular}} \\
                \midrule
                \multirow{3}{*}{\rotatebox[origin=c]{90}{Small}}
                & \multirow{2}{*}{\begin{tabular}[c]{@{}l@{}}Yago\\3-10\end{tabular}}
                & ComplEx & \textbf{0.536} & 0.463 & 0.528 & 0.552\\
                & & RotatE  & 0.432 & 0.432 & \textbf{0.434} & 0.453\tablefootnote{RotatE benefits from self-adversarial sampling as used in~\cite{sun2018rotate}. We did not use this technique to keep the search space consistent across all models. An adapted \textsc{GraSH} search space led to an MRR of 0.494 (combined, $d=512$), matching the prior result.}\setcounter{rotatenote}{\thefootnote}  \\
                & ($E=400$)& TransE  & \textbf{0.499} & 0.422 & \textbf{0.499} & 0.496 \\
                \midrule
                \multirow{3}{*}{\rotatebox[origin=c]{90}{Medium}}
                & \multirow{2}{*}{\begin{tabular}[c]{@{}l@{}}Wiki-\\data5M\end{tabular}}
                & ComplEx & \textbf{0.300} & \textbf{0.300} & \textbf{0.300} & 0.294 \\
                & & RotatE  & \textbf{0.241} & 0.232 & \textbf{0.241} & 0.261 \\
                & ($E=64$)& TransE  & 0.263 & 0.263 & \underline{\textbf{0.268}} & 0.249 \\
                \midrule
                \multirow{3}{*}{\rotatebox[origin=c]{90}{Large}}
                & \multirow{2}{*}{\begin{tabular}[c]{@{}l@{}}Free-\\base\end{tabular}}
                & ComplEx & 0.572 & \textbf{0.594} & \textbf{0.594} & \underline{0.678} \\
                & & RotatE  & 0.561 & \textbf{0.613} & \textbf{0.613} & \underline{0.615} \\
                & ($E=10$) & TransE  & 0.261 & \textbf{0.553} & \textbf{0.553} & \underline{0.559} \\
                \bottomrule
            \end{tabular}
            \label{tab:overall_performance}
        \end{subtable}
        \begin{subtable}[t]{.24\linewidth}
            \caption{Prior results}
            \label{tab:prior_results}
            \begin{tabular}{rrrr}
                \toprule
                \multirow{2}{*}{\begin{tabular}[r]{@{}c@{}}MRR\end{tabular}} &
                \multirow{2}{*}{\begin{tabular}[r]{@{}c@{}}Dim\end{tabular}} &
                \multirow{2}{*}{\begin{tabular}[r]{@{}c@{}}Epochs\end{tabular}} \\ \\
                \midrule
                \underline{\numprint{0.551}} & \numprint{128} & \numprint{400} & \cite{broscheit2020libkge}\tablefootnote{Published in the online appendix of~\cite{broscheit2020libkge}.} \\
                \underline{\numprint{0.495}}\footnotemark[\therotatenote]& \numprint{1000} & ? & \cite{sun2018rotate}\\
                \underline{\numprint{0.510}}\tablefootnote{Published with the AmpliGraph library~\cite{ampligraph}, which ignores unseen entities during evaluation. This inflates the MRR so that results are not directly comparable.} & \numprint{350} & \numprint{4000} & \cite{ampligraph} \\
                \midrule
                \underline{\numprint{0.308}} & \numprint{128} & \numprint{300} & \cite{kochsiek2021parallel}\\
                \underline{\numprint{0.290}} & \numprint{512} & \numprint{1000} & \cite{wang2019kepler}\\
                \numprint{0.253} & \numprint{512} & \numprint{1000} & \cite{wang2019kepler}\\
                \midrule
                \numprint{0.612} & \numprint{400} & \numprint{10} & \cite{kochsiek2021parallel} \\
                \numprint{0.567} & \numprint{128} & \numprint{10} & \cite{kochsiek2021parallel} \\
                - & - & - \\
                \bottomrule

            \end{tabular}
        \end{subtable}
        \label{tab:quality}
    \end{table}
}
\paragraphheader{Comparison to KGTuner.} KGTuner~\cite{zhang2022efficient} was
developed in parallel to this work and follows similar goals as \textsc{GraSH}.
We compared the two approaches on the smaller Yago3-10 KGE with ComplEx; a
comparison on the larger datasets was not feasible since KGTuner has large
computational costs. We ran both GraSH and KGTuner with the default settings of
KGTuner ($n=50$ trials, $E=50$ epochs, dim.~\numprint{1000}) to obtain a fair
comparison. KGTuner reached an MRR of $0.505$ in about 5 days (its search budget
corresponds to $B\approx20$). \textsc{GraSH} reached an MRR of $0.530$ in about $1.5$
hours ($B=3$, sequential search on 1 GPU), i.e., a higher quality result at
lower cost.
The high computational cost of KGTuner mainly stems from its inflexible and inefficient budget allocation (e.g., always 10 full-fidelity evaluations). The higher quality of \textsc{GraSH} stems from its use of multiple fidelities (vs.~two in KGTuner) and by using a combination of $k$-cores and epoch reduction (vs.~random walks in KGTuner).
\subsection{Resource Consumption (Tab.~\ref{tab:resource_consumption})}
\label{sec:runtime_comparison_exp}

{
\setlength{\tabcolsep}{1.84pt}
\begin{table}[t]
    \centering
    \caption[Resource consumption per round (ComplEx).]{Resource consumption per round (ComplEx).\footnotemark}
    \begin{tabular}{l@{\hskip .3in}lrrr@{\hskip 0.4in}rrr}
        \toprule
        & &
            \multicolumn{3}{c@{\hskip 0.4in}}{Round Time (min)} &
        \multicolumn{3}{c}{Model Size (MB)} \\
        & &
        \multicolumn{1}{c}{Epoch} &
        \multicolumn{1}{c}{Graph} &
                                    \multicolumn{1}{c@{\hskip 0.4in}}{Comb.} &
        \multicolumn{1}{c}{Epoch} &
        \multicolumn{1}{c}{Graph} &
        \multicolumn{1}{c}{Comb.} \\
        \midrule
        \multirow{4}{*}{Yago3-10}
        & Round 1 & \numprint{43.9} & \numprint{24.7} & \numprint{15.9} & \numprint{60.2} & \numprint{0.3} & \numprint{2.0} \\ %
        & Round 2 & \numprint{34.8} & \numprint{13.3} & \numprint{27.1} & \numprint{60.2} & \numprint{0.4} & \numprint{6.3} \\
        & Round 3 & \numprint{38.7} & \numprint{28.1} & \numprint{33.5} & \numprint{60.2} & \numprint{6.3} & \numprint{16.7} \\ %
        \cmidrule(rr{0.4in}){2-5}
        & Total & \numprint{117.4} & \textbf{\numprint{66.1}} & \numprint{76.5} \\
        \midrule
        \multirow{4}{*}{Wikidata5M}
        & Round 1 & \numprint{182.3} & \numprint{60.1} & \numprint{82.3} & \numprint{2353.3}   & \numprint{1.0}   & \numprint{71.3} \\
        & Round 2 & \numprint{134.2} & \numprint{87.4} & \numprint{88.6} & \numprint{2353.3}   & \numprint{36.0}  & \numprint{182.0} \\ %
        & Round 3 & \numprint{126.9} & \numprint{92.5} & \numprint{95.3} & \numprint{2353.3}   & \numprint{182.0} & \numprint{454.7} \\
        \cmidrule(rr{0.4in}){2-5}
        & Total & \numprint{443.4} & \textbf{\numprint{240.0}} & \numprint{266.2} \\
        \midrule
        \multirow{4}{*}{Freebase}
        & Round 1 & \numprint{915.9} & \numprint{250.7} & \numprint{179.7} & \numprint{42025.9}   & \numprint{87.3}   & \numprint{1322.2} \\
        & Round 2 & \numprint{507.9} & \numprint{172.0} & \numprint{151.2} & \numprint{42025.9}   & \numprint{520.1}  & \numprint{2667.7} \\
        & Round 3 & \numprint{423.4} & \numprint{197.5} & \numprint{207.0} & \numprint{42025.9}   & \numprint{2667.7} & \numprint{6571.3} \\
        \cmidrule(rr{0.4in}){2-5}
        & Total & \numprint{1847.2} & \numprint{620.2} & \textbf{\numprint{537.9}}\\
        \bottomrule
    \end{tabular}
    \label{tab:resource_consumption}
\end{table}
}

\footnotetext{The time needed to compute the $k$-core decompositions is excluded.
  It is negligible compared to the overall search time (e.g., $\approx28\,$min for Freebase with igraph~\cite{csardi2006igraph}).}

Next, we investigated the computational cost and memory consumption of each
round of \textsc{GraSH}. We used 4 GPUs in parallel, evaluating one trial per
GPU with the same settings as used in
Sec.~\ref{sec:model_quality}. %
Our results are summarized in Tab.~\ref{tab:resource_consumption}.

\paragraphheader{Memory consumption.} Epoch reduction was less effective than
graph reduction and a combined approach in terms of memory usage. With epoch
reduction, training is performed on the full graph in every round and therefore
performed with full model size. Due to the large model sizes on the largest
graph Freebase, the model could not be kept in GPU memory introducing further
overheads for parameter management.
Graph reduction with $k$-core decomposition reduced the number of entities
contained in a subgraph considerably. As the model size is mainly driven by the
number of entities, the resulting model sizes were small.

\paragraphheader{Runtime.} Similarly to memory consumption, a \textsc{GraSH}
search based on epoch reduction was less effective in terms of runtime compared
to graph reduction and a combined approach. With epoch reduction, runtime was
mainly driven by the cost of model evaluation and model initialization (see
Sec.~\ref{sec:background} and~\ref{sec:epoch_reduction}). This is especially
visible in the first round of the search on large graphs. Here, the number of
trials and therefore the number of model initializations and evaluations is
high. Additionally, on the largest graph, the overhead for parameter management
for training on the full KG increased runtime further. In contrast, small model
sizes and low GPU utilization with graph reduction would even allow further
performance gains. For example, improving on the presented results, the runtime
of the first round on Wikidata5M can be reduced from $60.1$ to $22.9$ minutes by
training three models per GPU instead of one.

\subsection{Influence of Number of Rounds (Tab.~\ref{tab:eta_influence})}
\label{sec:robustness}

{
\setlength{\tabcolsep}{3.9pt}
\begin{table}[t]
    \centering
    \caption{Influence of the number of rounds on model quality in terms of MRR
      (ComplEx, graph reduction, $n=64$ trials, $B=3$). The number of rounds is directly controlled by the choice of $n$ and $\eta$.}
    \begin{tabular}{lccc|cc}
        \toprule
        \multicolumn{1}{l}{} &
        \multicolumn{1}{c}{$\eta=2$} &
        \multicolumn{1}{c}{$\eta=4$} &
        \multicolumn{1}{c|}{$\eta=8$} &
        \multicolumn{1}{c}{$\eta=64$} &
        \multicolumn{1}{c}{$\eta=64$} \\
        Dataset & 6 rounds & 3 rounds & 2 rounds & 1 round & 1 round \\
        & & (default) & & $B=3$ & $B=1$\\
        \midrule
        Yago3-10 & 0.463  & 0.463 & 0.485 & 0.427 & 0.427 \\
        Wikidata5M & 0.300 & 0.300 & 0.300 & 0.300 & 0.285 \\
        Freebase & 0.594 & 0.594 & 0.594 & 0.572 & 0.572 \\
        \bottomrule
    \end{tabular}
    \label{tab:eta_influence}
\end{table}
}

In our final experiment, we investigated the sensitivity of \textsc{GraSH} with
respect to the number of rounds being used as well as whether using
multi-fidelity optimization is beneficial. Our results are summarized in
Tab.~\ref{tab:eta_influence}. All experiments were conducted at the same budget
($B=3$) and number of trials ($n=64$). Note that the number of rounds used by
\textsc{GraSH} is given by $\log_\eta(n)$, where $n$ denotes the number of trials
and $\eta$ the reduction factor. The smaller $\eta$, the more rounds are used and the
lower the (initial) fidelity.

We found that on the two larger graphs, the search was robust to changes in
budget allocation and $\eta$ did not influence the final trial selection (as long
as at least 2 rounds were used). Only on the smaller Yago3\nobreakdash-10 KG,
the final model quality differed with varying values of $\eta$. Here, low-fidelity
approximation (small~$\eta$) was riskier since the subgraphs used in the first rounds were very small.

To investigate whether multi-fidelity HPO---i.e., multiple rounds---are beneficial,
we (i) used the best configuration of the first round directly ($\eta=64$, $B=1$) and
(ii) performed an additional single-round search with a comparable budget to all
other settings ($\eta=64$, $B=3$). As shown in Tab.~\ref{tab:eta_influence}, both settings did not reach the performance
achieved via multiple rounds. We conclude that the use of multiple fidelity
levels is essential for cost-effective HPO.

\section{Conclusion}
\label{sec:conclusion}
We first presented and experimentally explored various low-fidelity
approximation techniques for evaluating hyperparameters of KGE models. Based on
our findings, we proposed \textsc{GraSH}, an open-source, multi-fidelity
hyperparameter optimizer for KGE models based on successive halving. We found
that \textsc{GraSH} often reproduced or outperformed state-of-the-art results on
large knowledge graphs at very low overall cost, i.e., the cost of three
complete training runs. We argued that the choice of low-fidelity approximation
is crucial ($k$-core reduction combined with epoch reduction worked best), as is
the use of multiple fidelities.

\bibliographystyle{splncs04}
%
\bibliography{main}

\begin{thebibliography}{10}
\providecommand{\url}[1]{\texttt{#1}}
\providecommand{\urlprefix}{URL }
\providecommand{\doi}[1]{https://doi.org/#1}

\bibitem{ali2021bringing}
Ali, M., Berrendorf, M., Hoyt, C.T., Vermue, L., Galkin, M., Sharifzadeh, S.,
  Fischer, A., Tresp, V., Lehmann, J.: Bringing light into the dark: A
  large-scale evaluation of knowledge graph embedding models under a unified
  framework. IEEE Transactions on Pattern Analysis and Machine Intelligence
  (2021)

\bibitem{baier2017improving}
Baier, S., Ma, Y., Tresp, V.: Improving visual relationship detection using
  semantic modeling of scene descriptions. In: International Semantic Web
  Conference. pp. 53--68. Springer (2017)

\bibitem{baker2017accelerating}
Baker, B., Gupta, O., Raskar, R., Naik, N.: Accelerating neural architecture
  search using performance prediction. In: International Conference on Learning
  Representations (Workshop) (2018)

\bibitem{bordes2013translating}
Bordes, A., Usunier, N., Garcia-Duran, A., Weston, J., Yakhnenko, O.:
  Translating embeddings for modeling multi-relational data. Advances in neural
  information processing systems  \textbf{26},  2787--2795 (2013)

\bibitem{broscheit2020libkge}
Broscheit, S., Ruffinelli, D., Kochsiek, A., Betz, P., Gemulla, R.: Libkge a
  knowledge graph embedding library for reproducible research. In: Proceedings
  of the 2020 Conference on Empirical Methods in Natural Language Processing
  (2020)

\bibitem{ampligraph}
Costabello, L., Pai, S., Van, C.L., McGrath, R., McCarthy, N., Tabacof, P.:
  {AmpliGraph: a Library for Representation Learning on Knowledge Graphs} (Mar
  2019)

\bibitem{csardi2006igraph}
Csardi, G., Nepusz, T., et~al.: The igraph software package for complex network
  research. InterJournal, complex systems  \textbf{1695}(5), ~1--9 (2006)

\bibitem{ConvE}
Dettmers, T., Minervini, P., Stenetorp, P., Riedel, S.: Convolutional 2d
  knowledge graph embeddings. In: Proceedings of the 32nd {AAAI} Conference on
  Artificial Intelligence. pp. 1811--1818 (2018)

\bibitem{falkner2018bohb}
Falkner, S., Klein, A., Hutter, F.: Bohb: Robust and efficient hyperparameter
  optimization at scale. In: International Conference on Machine Learning. pp.
  1437--1446. PMLR (2018)

\bibitem{jamieson2016non}
Jamieson, K., Talwalkar, A.: Non-stochastic best arm identification and
  hyperparameter optimization. In: Artificial intelligence and statistics. pp.
  240--248. PMLR (2016)

\bibitem{kochsiek2021parallel}
Kochsiek, A., Gemulla, R.: Parallel training of knowledge graph embedding
  models: a comparison of techniques. Proceedings of the VLDB Endowment
  \textbf{15}(3),  633--645 (2021)

\bibitem{lacroix2018canonical}
Lacroix, T., Usunier, N., Obozinski, G.: Canonical tensor decomposition for
  knowledge base completion. In: Proceedings of 35th International Conference
  on Machine Learning. pp. 2863--2872. PMLR (2018)

\bibitem{lerer2019pytorch}
Lerer, A., Wu, L., Shen, J., Lacroix, T., Wehrstedt, L., Bose, A.,
  Peysakhovich, A.: Pytorch-biggraph: A large-scale graph embedding system.
  Proceedings of the 2nd SysML Conference  (2019)

\bibitem{mohamed2019drug}
Mohamed, S.K., Nounu, A., Nov{\'a}{\v{c}}ek, V.: Drug target discovery using
  knowledge graph embeddings. In: Proceedings of the 34th ACM/SIGAPP Symposium
  on Applied Computing. pp. 11--18 (2019)

\bibitem{nickel2015review}
Nickel, M., Murphy, K., Tresp, V., Gabrilovich, E.: A review of relational
  machine learning for knowledge graphs. Proceedings of the {IEEE}  (2015)

\bibitem{nickel2011three}
Nickel, M., Tresp, V., Kriegel, H.P.: A three-way model for collective learning
  on multi-relational data. In: Proceedings of the 28th International
  Conference on Machine Learning. vol.~11, pp. 809--816 (2011)

\bibitem{ruffinelli2020you}
Ruffinelli, D., Broscheit, S., Gemulla, R.: You {CAN} teach an old dog new
  tricks! on training knowledge graph embeddings. In: International Conference
  on Learning Representations (2020)

\bibitem{saxena2022sequence}
Saxena, A., Kochsiek, A., Gemulla, R.: Sequence-to-sequence knowledge graph
  completion and question answering. In: Proceedings of the 60th Annual Meeting
  of the Association for Computational Linguistics (Volume 1: Long Papers). pp.
  2814--2828 (2022)

\bibitem{saxena2020improving}
Saxena, A., Tripathi, A., Talukdar, P.: Improving multi-hop question answering
  over knowledge graphs using knowledge base embeddings. In: Proceedings of the
  58th annual meeting of the association for computational linguistics. pp.
  4498--4507 (2020)

\bibitem{seidman1983network}
Seidman, S.B.: Network structure and minimum degree. Social networks
  \textbf{5}(3),  269--287 (1983)

\bibitem{sun2018rotate}
Sun, Z., Deng, Z.H., Nie, J.Y., Tang, J.: Rotate: Knowledge graph embedding by
  relational rotation in complex space. In: International Conference on
  Learning Representations (2019)

\bibitem{toutanova2015observed}
Toutanova, K., Chen, D.: Observed versus latent features for knowledge base and
  text inference. In: Proceedings of the 3rd workshop on continuous vector
  space models and their compositionality. pp. 57--66 (2015)

\bibitem{Complex}
Trouillon, T., Welbl, J., Riedel, S., Gaussier, {\'E}., Bouchard, G.: Complex
  embeddings for simple link prediction. In: International conference on
  machine learning. pp. 2071--2080. PMLR (2016)

\bibitem{tu2019autone}
Tu, K., Ma, J., Cui, P., Pei, J., Zhu, W.: Autone: Hyperparameter optimization
  for massive network embedding. In: Proceedings of the 25th ACM SIGKDD
  International Conference on Knowledge Discovery \& Data Mining. pp. 216--225
  (2019)

\bibitem{wang2017knowledge}
Wang, Q., Mao, Z., Wang, B., Guo, L.: Knowledge graph embedding: A survey of
  approaches and applications. IEEE Transactions on Knowledge and Data
  Engineering  \textbf{29}(12),  2724--2743 (2017)

\bibitem{wang2021rank}
Wang, R., Chen, X., Cheng, M., Tang, X., Hsieh, C.J.: Rank-nosh: Efficient
  predictor-based architecture search via non-uniform successive halving. In:
  Proceedings of the IEEE/CVF International Conference on Computer Vision. pp.
  10377--10386 (2021)

\bibitem{wang2019kepler}
Wang, X., Gao, T., Zhu, Z., Liu, Z., Li, J., Tang, J.: Kepler: A unified model
  for knowledge embedding and pre-trained language representation. Transactions
  of the Association for Computational Linguistics  (2021)

\bibitem{yang2015embedding}
Yang, B., Yih, S.W.t., He, X., Gao, J., Deng, L.: Embedding entities and
  relations for learning and inference in knowledge bases. In: Proceedings of
  the International Conference on Learning Representations (2015)

\bibitem{zhang2022efficient}
Zhang, Y., Zhou, Z., Yao, Q., Li, Y.: Efficient hyper-parameter search for
  knowledge graph embedding. In: Proceedings of the 60th Annual Meeting of the
  Association for Computational Linguistics (Volume 1: Long Papers). pp.
  2715--2735 (2022)

\bibitem{Zheng2020DGLKETK}
Zheng, D., Song, X., Ma, C., Tan, Z., Ye, Z.H., Dong, J., Xiong, H., Zhang, Z.,
  Karypis, G.: Dgl-ke: Training knowledge graph embeddings at scale.
  Proceedings of the 43rd International ACM SIGIR Conference on Research and
  Development in Information Retrieval  (2020)

\bibitem{zwillinger1999crc}
Zwillinger, D., Kokoska, S.: CRC standard probability and statistics tables and
  formulae. Crc Press (1999)

\end{thebibliography}

\end{document}